\documentclass[letterpaper, 10pt, conference]{ieeeconf}
\IEEEoverridecommandlockouts

\usepackage[pdftex]{graphicx}
\usepackage{multicol}
\usepackage{amsmath, amsfonts, amssymb}
\usepackage{xspace}
\usepackage{latexsym}
\usepackage{url}
\usepackage{listings}
\usepackage{float}
\usepackage{tabularx}
\usepackage{tikz}
\usepackage{fancybox}
\usepackage{flushend}
\usepackage{tabulary}
\usepackage{lipsum}
\usepackage[ruled,lined]{algorithm2e}
\usepackage[utf8]{inputenc}
\usepackage{fourier} 
\usepackage{array}
\usepackage{makecell}
\usepackage{colortbl}
\usepackage{xcolor}
\usepackage[export]{adjustbox}
\usepackage[hidelinks]{hyperref}
\usepackage{comment}

\definecolor{Gray}{gray}{0.85}

\lstdefinestyle{Prolog} {
    language=Prolog, 
    lineskip=0.2ex, 
    fontadjust=true, 
    basicstyle={\scriptsize \nopagebreak[4]}, 
    commentstyle=\scriptsize,
    morekeywords={entity, occurs, holds, show, append, forall, findall, member, type}
    }

\newcommand{\knowrob}{\textsc{KnowRob}}

\newcommand{\openease}{\textsc{openEASE}}
\newcommand{\owl}{\textsc{OWL}}

\graphicspath{{img/}}

\title{Automated acquisition of structured, semantic models of
  manipulation activities from human VR demonstration}

\author{Andrei Haidu and Michael Beetz\\
    {\{haidu, beetz\}@uni-bremen.de}
    
    \thanks{The authors are with the Institute for Artificial Intelligence, 
    University of Bremen, Germany.} 
    \thanks{This work was partially funded by Deutsche Forschungsgemeinschaft
    (DFG) through the Collaborative Research Center 1320, EASE, and by the EU H2020 project \emph{REFILLS} (Project ID: 731590).}
}

\begin{document}

This work has been submitted to the IEEE for possible publication.
Copyright may be transferred without notice, after which this version
may no longer be accessible.
\newpage

\maketitle

\begin{abstract}

 In this paper we present a system capable of collecting and annotating,
 human performed, robot understandable, everyday activities from virtual
 environments. The human movements are mapped in the simulated world
 using off-the-shelf virtual reality devices with full body, and eye
 tracking capabilities. All the interactions in the virtual world are
 physically simulated, thus movements and their effects are closely
 relatable to the real world. During the activity execution, 
 a \emph{subsymbolic data} logger is recording the environment and
 the human gaze on a per-frame basis, enabling offline scene reproduction
 and replays. Coupled with the physics engine, online monitors
 (\emph{symbolic data} loggers) are parsing (using various grammars)
 and recording events, actions, and their effects in the simulated world.
 

\end{abstract}

\section{Introduction}
\label{sec:intro}

We are currently seeing fast progress of robotic agents learning
manipulation tasks \cite{DeepManipRL}, in particular through imitation
\cite{billard08robot} and reinforcement learning \cite{RoboticsRL}.
However, in many cases the amount of experience
needed for learning exceeds what can provided with reasonable effort
\cite{google-pick-ijrr18}, in particular if a single experience corresponds
to a complete manipulation episode. One way of reducing the number of
experiences needed for learning is to increase the information content
of an episode. This can be achieved by transforming experiences into
appropriately structured activity models and interpreting episodes by
applying structural and teleological, as well as causal and intuitive
physics knowledge to them. While several approaches are starting to
look towards this direction \cite{ramirez17} we believe that there is
much higher potential in such activity representations as human
activities are highly structured and have many regularities
\cite{flanagan06control,land06} that have been investigated thoroughly in an
interdisciplinary research field called action science \cite{prinz13}.

In this paper we advance the acquisition of learning data from
human virtual reality demonstrations by building on top of our previous
work: AMEvA (Automated Models of Everyday Activities)
\cite{haidu19ameva,haidu18gameengine} -- a special-purpose knowledge
acquisition, interpretation and processing framework, capable
of recording and detecting force-dynamic states of
manipulation activities performed in virtual environments.
The main limitations of the previous system were the coarser
representations of the human models and their actions.
The human model was consisting of only the head and hands, tracked
by the VR headset and the two controllers. This approximation implies
that we can only get trajectory data of the hands 
and the object manipulated by the hand but not full-pose data of 
the human body. Also, the model allows the human to reach locations 
that are impossible to reach in reality because other body parts 
would collide with the environment, or cannot use various body 
parts to manipulate the environment, for example closing a drawer 
or a door using an elbow or a foot.

\begin{figure}[t]
    \begin{center}
        \includegraphics[width=\columnwidth]%
        {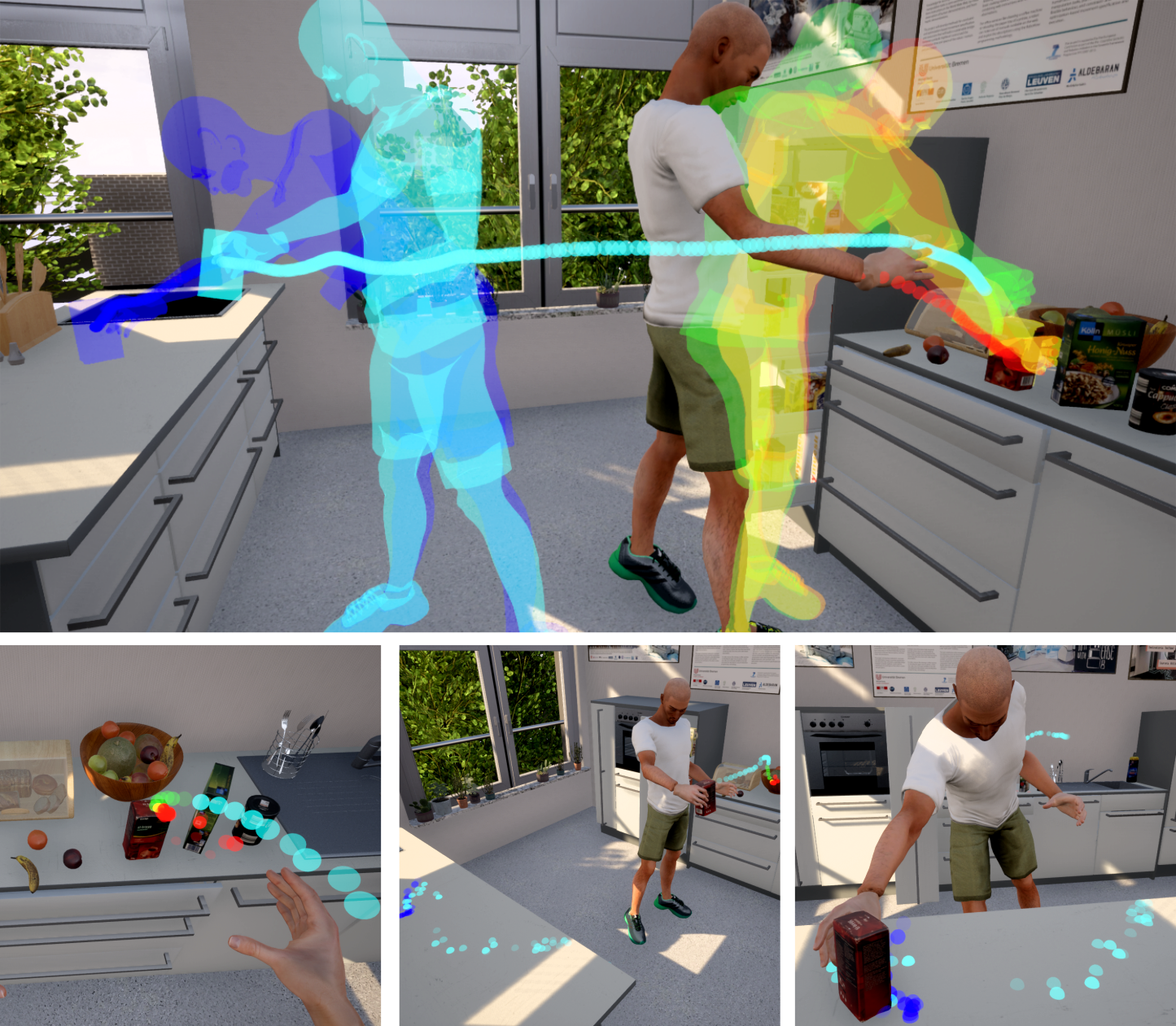}
        \caption{Pick-and-place motion patterns with gaze visualization}
        \vspace*{-0.85cm}
        \label{fig:main} 
    \end{center}
\end{figure}

\begin{figure*}[!t]
    \centering 
    \includegraphics[width=\textwidth]%
    {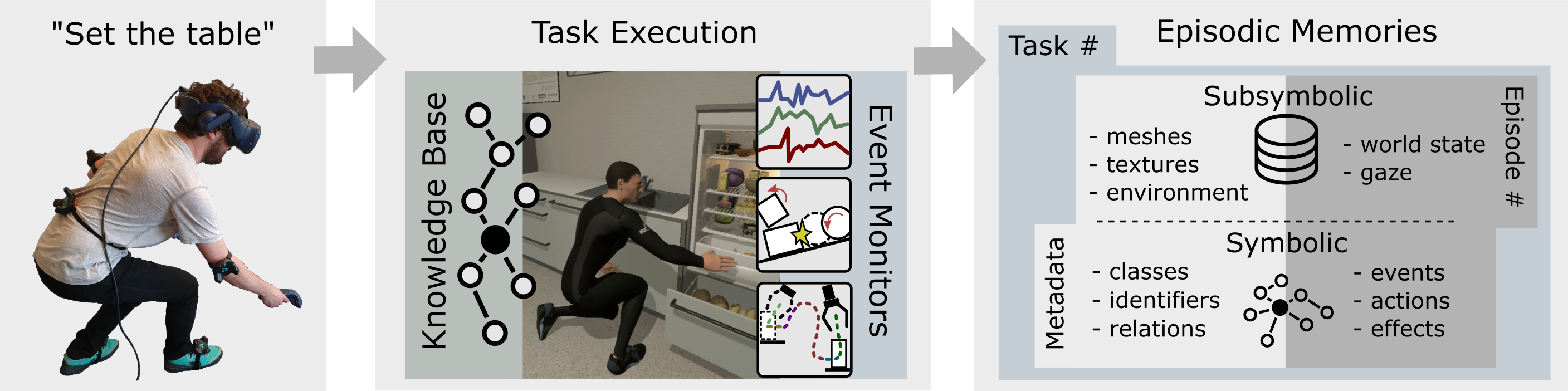}
    \caption{System overview}
    \vspace*{-0.5cm}
    \label{fig:overview}
\end{figure*}

As a result, the contributions of this paper are: \emph{(a)} the improved
human model representation; \emph{(b)} a more realistic interaction
with the virtual world; and \emph{(c)} the granularity of the action 
and event recognition/segmentation; For \emph{(a)}, we integrated gaze
and full body tracking systems to provide us in depth access to the
human motions and intentions. The virtual human body obtained an in-depth
semantic representation for its skeletal body parts using \knowrob\ 
\cite{beetz18knowrob}, thus commonly shared with the robot. This was done
by extending the upper ontology with classes, properties and relations
of the virtual body parts. In \emph{(b)} the primary interaction 
advancement is a fully physics-based grasping model, where grasping
occurs as a result of opposing forces and friction applied to an object
through the hand parts. The integrated full body allows, as previously
mentioned, for more natural and complex interactions, such as closing
doors/drawers using other body parts than the hands (elbows, foot).
For \emph{(c)}, the main contribution is the implementation of more
granular online monitors capable of detecting and segmenting generic,
physics-enabled interactions. We extended the Flanagan model
\cite{flanagan2006control} for pick-and-place being able to robustly
detect reach-to-grasp (prehension) movements such as \emph{reaching},
\emph{fixation}, \emph{grasping}, \emph{sliding}, \emph{picking-up},
\emph{transporting}, \emph{putting-down}.

Figure~\ref{fig:main} partially illustrates the aforementioned
contributions by depicting a segmented pick-and-place action 
into meaningful motion patterns (top) and the gaze of the human
during the same execution (bottom).
The scene is reproduced at the start of the 
reaching motion and annotated with colored markers 
(hand trajectory and human pose) for each following motion phase.
Red for reaching, yellow for fixation, green for picking-up,
teal for transporting and blue for the putting-down motion.

The remainder of the paper is organized as follows: 
Section~\ref{sec:overview} gives an overview of the framework,
describes the symbolic virtual environment and the structure 
of the episodic memories connected with the robot's knowledge base.
Section~\ref{sec:interaction} presents the interaction 
of the human with the virtual environment. 
Section~\ref{sec:activity_parser} introduces the 
online activity recognition modules and their heuristics.
Section~\ref{sec:experiments} introduces example queries for
accessing the experiments data from the hybrid knowledge base.
We introduce various related works in section~\ref{sec:related-work},
and conclude in section~\ref{sec:conclusion}.

\section{Overview}
\label{sec:overview}

Figure~\ref{fig:overview} depicts the overview of the system,
where a user is asked to perform a task in a virtual environment.
The environment is connected to an extended \knowrob\ knowledge base,
which includes the representation of all the entities and their properties
in the virtual world (humans, objects, lights, articulations, partonomies),
as well as the actions and events which can occur during the task execution. 
A \emph{subsymbolic}, and \emph{symbolic}, logger is running in the
background, recording on a per frame basis every change in the environment
(including human gaze), where the latter is a collection of modules
capable of detecting and recording actions and events. These are then 
indexed and stored in a hybrid knowledge base in the form 
of episodic memories.

\subsection{Virtual Environment}

The data structure used in the virtual world can be split into
two categories: ``materialized'' and ``abstract'' entities. These
can further be categorized as relevant for the rendering engine,
the physics engine, for both, or for none.
The ``materialized'' entities are the ones that can be
drawn by the rendering engine \emph{and} simulated by the physics engine:
the human body (skeletal body), the objects in the environment
(rigid bodies) or physics-based particles such as liquids or soft bodies.
``Abstract'' entities, such as lights or visual effects particles
(sparks, flames etc.) are only relevant for the rendering engine,
whereas joint/spring entities are only relevant for the physics engine.
Custom abstract entities such as gaze pose are irrelevant for both,
since they are not simulated nor rendered. Depending on their types,
the entities have various properties, such as collisions, mass,
friction parameters, textures, etc. for ``materialized'' entities.
Emitted light color, intensity, shape, etc. for lights. Motion limits
(prismatic/linear), damping, stiffness coefficients, etc. for joints.
Another important property is the partonomy relationships between
the entities, for example that the drawer handle is part of the drawer.
For each entity and property there is a corresponding representation
in the extended ontology of \knowrob.

\subsection{Episodic Memories}
\label{sec:ep_mem}

The right side of Figure~\ref{fig:overview} depicts the structure of the
logged datasets. It is split into a symbolic and subsymbolic 
representation. The symbolic part is stored in \knowrob\ and represented
using \owl~\cite{Horrocks03}, whereas for the subsymbolic part we are
using MongoDB databases. Examining the dataset from top to bottom,
we first have a number of \emph{tasks} stored using a \knowrob\ instance
bundled with a MongoDB server. A task represents a virtual environment
scenario where users are given a specific assignment. Tasks are
concepts in the ontology, where each new task is represented as a unique
instance in the knowledge base and a separate \texttt{database} in MongoDB.
Such a task example would be to: "set the table for dinner",
"clean the table after breakfast", or "put the dirty dishes in the
dishwasher". Each task component is split into two parts: the
\emph{metadata}, which stores the data at the task level, and the 
\emph{episodes} collection, which are the instances of every executed 
tasks. The concept of ``episode'' is also represented in the ontology,
therefore every episode instance will receive a unique identifier at 
instantiation, thus allowing \knowrob\ to classify each occurred event
per episode and to query the knowledge base as a whole.

The symbolic section of the \emph{metadata} represents the instantiation
of every entity class from the virtual world, which entails: generating
unique identifiers using \texttt{UUID}'s for each entity, 
thus guaranteeing uniqueness of all the instances across any
previous or future tasks;
and collecting all properties of the entities.
The subsymbolic section stores all the binary assets required for the 
virtual environment to re-build the task scenario: meshes, materials,
textures, shaders, lighting builds of the map, etc. These are stored 
using MongoDB's large files management system \texttt{gridfs}.

At the symbolic level each \emph{episode} stores all the events,
actions and their effects that occurred during the task execution,
these are stored into per episode separate \owl\ files.
At the subsymbolic level, for each episode a new MongoDB
\texttt{collection} is created, storing the world states on a per-frame
basis and the gaze information.
Since the database is no longer altered after the episode completion,
this allows us to optimize the database for fast querying using
multiple \texttt{indexes}. Every relevant field will be indexed 
and internally organized as a \texttt{B-tree} data structure,
thus reducing the search time complexity to $\mathcal{O}(\log{}n)$.

\section{Interaction}
\label{sec:interaction}

In this section we present how the mapping of the human movements 
onto the virtual avatar is implemented. In order to interact with the
environment we are using three types of interactions. First, we have the
full body tracking data applied to the virtual human body as an animation.
Second, a more refined controller is used for the hands movement,
allowing for a more precise fine-tuning to approximate real world
conditions. Finally, we have a physics-based grasp system capable
of switching between various predefined grasp styles. With the three
interactions styles the user gains full control over its virtual movements,
giving him/her the freedom to move in the open world as naturally
as possible.

\subsection{Full Body Tracking}

The full body tracking system is currently a live animation, meaning 
it will map every received bone position from the tracking software to
the avatar, neglecting the environment. Basically, it can apply forces
to the environment, but it cannot receive any feedback. Thus, if willing,
the user could cause instabilities in the simulated world by walking
through walls or furniture. Although the limitations, the user can still
take advantage of the additional limbs to interact with the world,
and will force him/her to use natural movements when doing so.

\subsection{Hand Control}

In order to have a more close-to-reality interaction, we needed to use
a different approach for the parts predominantly used to interact with
the world, namely the hands. A direct mapping would have made the
interaction most reactive, it however would also cause the user to interact
with objects as they would have been weightless. We therefore implemented
an approach to allow a two-way interaction with the world. When lifting a
heavy object, the object would pull back on the hands, or slip out.
This would compel the user to use strategies to interact with the
environment, as it would in the real world, such as using two hands
for stabilizing objects, or for lifting heavier ones.
For the implementation we used two 3D \texttt{PID} controllers, one 
for the translational motion and one for the rotational one. 
The controllers are in a closed loop with the tracking system, from which
they receive the 6D target pose and then apply forces and torques
to the hand to try to move it to the desired pose. The controller 
parameters gives us the possibility to calibrate the hands in order to
match close to real world values in terms of responsiveness 
and strength.

\subsection{Physics-based Grasping}
\label{sec:grasping}

\begin{figure}[htb]
    \begin{center}		
        \includegraphics[width=\columnwidth]%
        {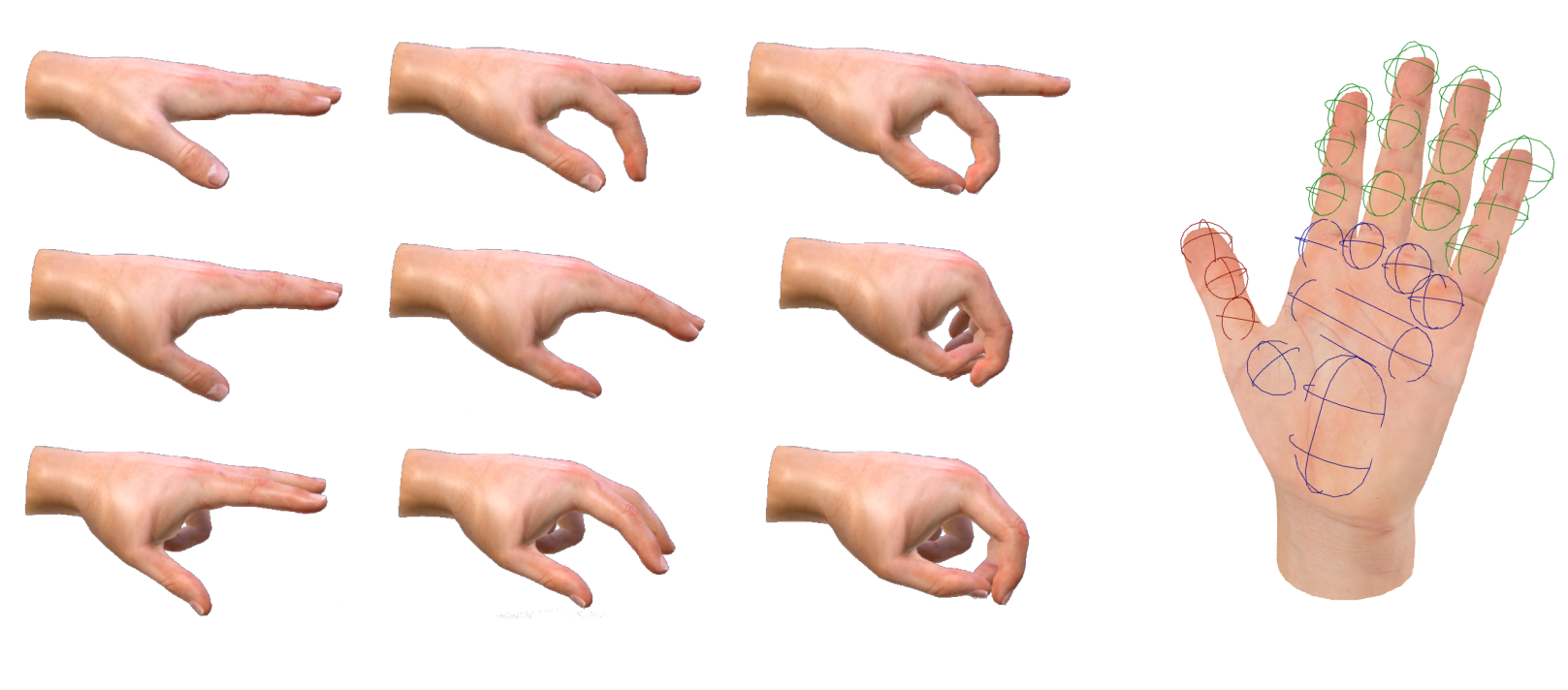}
        \caption{Physics-based grasp animations (left) and contact sensors (right)}
        \vspace*{-0.25cm}
        \label{fig:grasp_anim} 
    \end{center}
\end{figure}

In order to have plausible action effects during the task execution 
we implemented a physics-based grasping model on top of the simulated
hands. We have chosen the most common grasp styles for everyday object
manipulation and classified them according to Feix \textit{et al.}~\cite{GRASP2016}.
The grasp styles were created using standard 3d animation tools,
where each grasp consists of multiple frames storing the joint state
information of the skeletal hand.
The left part of Figure~\ref{fig:grasp_anim} shows an example of three 
grasp styles (top to bottom: pinch, wrap and tripod) and some of their
frames as waypoints. These animations are then stored in into
``physics grasp animation'' datasets which are loaded by the 
grasp controller. The controller will therefore react to two user inputs:
a discrete one, which changes the grasp style; and an analog one,
which interpolates the button mapping between the grasp animation frames,
resulting in the target angle for the joints.

To control the bone movements, each joint is equipped with a driver,
which takes as input the interpolated value from the grasp animation.
Torques are then applied to move and maintain the joint at the given
angle. The drivers operate similar to a PD controller using two inputs:
stiffness and damping. Where stiffness controls the strength of the
drive towards the target joint angle, and damping towards the target
velocity. In other words the strength of executing the grasp and the
strength of maintaining the grasp pose from any external perturbation.

\begin{figure}[htb]
    \begin{center}		
        \includegraphics[width=\columnwidth]%
        {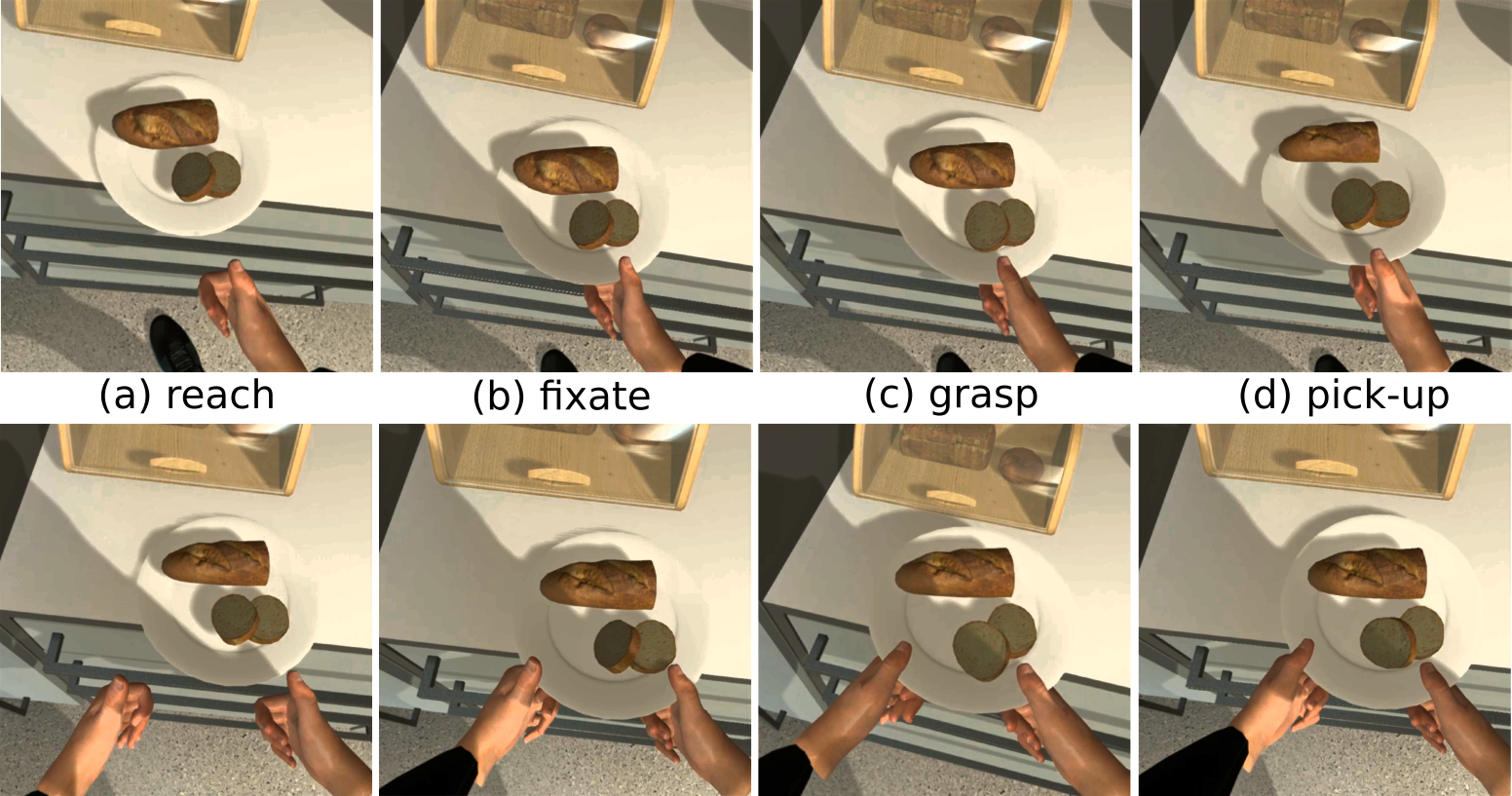}
        \caption{Na\"ive physics reasoning capabilities}
        \vspace*{-0.5cm}
        \label{fig:thumb_grasp_and_sensors} 
    \end{center}
\end{figure}

In Figure~\ref{fig:thumb_grasp_and_sensors} we showcase the real world
plausibility of the physics-based interaction systems. We depict a
timeframe of picking up a filled plate using a lateral grasp style
with one or two hands. We notice on the top timeframe during the pick-up
motion due to its weight the plate twists out of the grasp. However,
using two hands, the weight is distributed among two points, thus greatly
reducing the twist force applied by the plate, resulting in a successful
pick-up. Such events can provide robots with na\"ive physics understanding
of their actions. Having the grasp styles represented as concepts
in the ontology, it will also provide the robot with a mapping between
grasp styles and objects, or actions.

\section{Online Activity Parser}
\label{sec:activity_parser}

In the following section we present the online activity and event
detection modules running in realtime during the task execution
and symbolically record the various events occurring in the virtual world.

\subsection{Grasp Detection}

As described in the previous subsection \ref{sec:grasping} 
the current grasping model is physics-based, therefore, even though
we have the user's intention to grasp something, the actual success of
the action depends on the surrounding environment. In order to
differentiate between a contact with the hand while grasping and an
actual grasp event, we equip the virtual hands with arrays of contact
sensors (see right side of Figure~\ref{fig:thumb_grasp_and_sensors})
to detect if the user has something in the hand while executing the grasp
animation. The contact arrays are grouped into different sets, in the
example image we have green for the finger bones (phalanges), red for the
thumb, and blue for the palm (carpal bones). The grasp detector module
is only activated when the user starts the grasp animation. The module
is then keeping track of all the objects in contact with the sensors in
the sets. A grasp is detected when an object is in contact with sensors
from at least two different sets. Having the sensors covering
most of the hand area, it can can also detect grasping multiple
objects simultaneously.

The module will not differentiate between the grasping contexts,
for example, grasping an object to be transported, a handle to open
a drawer, or grasping an unmovable object such as the furniture.
It will regardlessly semantically annotate, and record the occurred grasping
event containing: the start and end time of the event, the hand and
the grasped object instances, including the grasp style used for
performing the action. Using \knowrob\ these events can further be
particularized, using first order logic rules to infer grasps related
events such as ``grasping onto'' or ``holding onto'' if during the
grasping event the ``acted on'' instance has a mass larger than
the strength of what the grasping hand can handle.

The grasp detection system is susceptible for different grammars
for detecting grasps, it can handle more than three sets of sensors array
if required. For examples if one would like to detect lateral finger
grasps (these however are very uncommon in a kitchen environment), one
could add for each finger a different set of sensor array.

\begin{figure*}[htb]
    \centering 
    \includegraphics[width=\textwidth]%
    {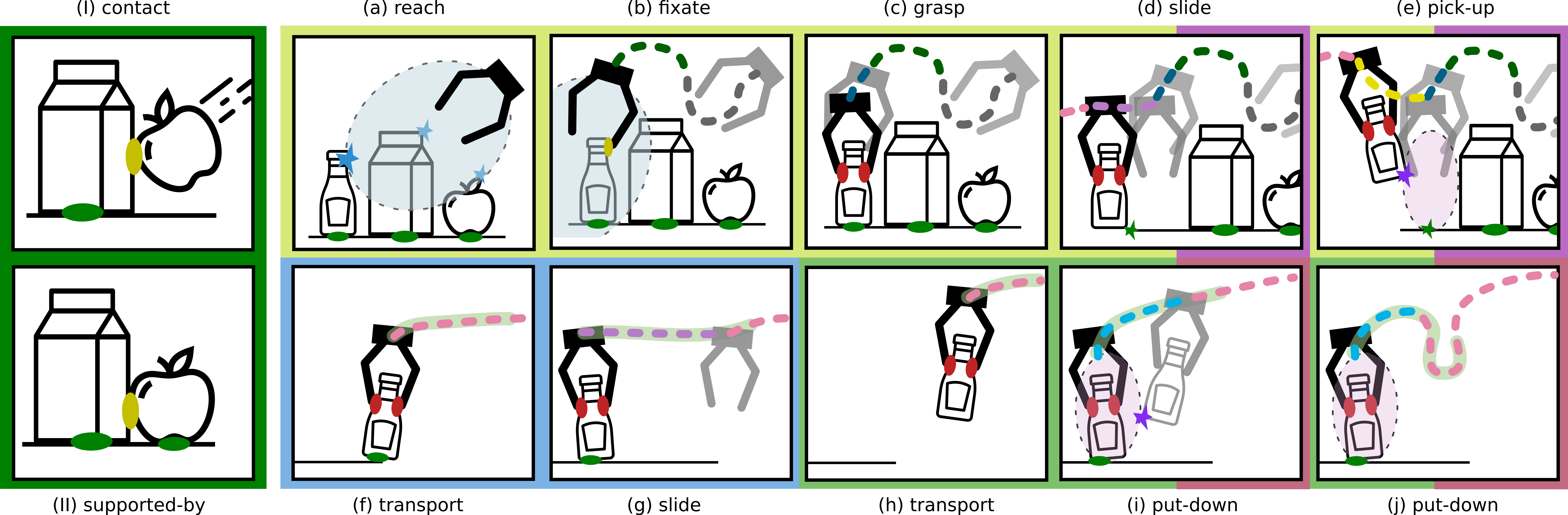}
    \caption{Key frames of the activity parsing grammar}
    \vspace*{-0.5cm}
    \label{fig:pick-and-place} 
\end{figure*}

\subsection{Contact and Supported-by} 

One of the building blocks for detecting force-dynamic events are the
\emph{contact} and \emph{supported-by} events \cite{fern02}. In our 
framework the two are bundled together in one module which is equipped 
on every object in the world. Tightly coupled with the physics engine,
the module will subscribes for any collisions occurring on the object.
When a collision is triggered, it is interpreted into a semantic contact 
event, and it starts checking for supported-by event until the contact
finishes. The grammar for checking for the supported-by event requires
that the two objects in contact have a relative vertical velocity of
close to zero. This way if an object is being transported
on a tray will still be detected as \emph{supported-by} the tray.

Being tightly coupled with the physics engine, the modules are only
active when the objects are simulated. For optimization reasons the
physics engine disables all objects that are in a stable state, when
this happens, the modules are also deactivated. This way the system can
scale to support thousands of objects without effecting the real-time
performance. Once an object is being simulated again, the corresponding
module is re-activated and continues listening for events.

In the left side of Figure~\ref{fig:pick-and-place} we show an example
of detecting contact and supported-by events between two objects and
a supporting surface. The yellow ellipsoid areas represent contact events,
whereas the green ones supported-by.

\subsection{Extended Pick-and-Place Detection}

Building upon the aforementioned modules, we now introduce the grammar
used for detecting and segmenting a pick-and-place action into relevant
motion patterns. Figure~\ref{fig:pick-and-place} depicts the most
significant stages used by the grammar to detect the 
reach-to-grasp action (top) and transport with placing (bottom).

For every pick-and-place action we first start detecting the 
\emph{reaching} motion, which is an important motion pattern for any robot
control system executing manipulations tasks. In our model, reaching
represents: the motion of the user's hand towards an object which will
eventually be grasped. The reaching motion is finished when the hand
(contact sensors) are in contact with the to-be-grasped object. The
starting point for reaching is hard to detect because there are no
distinct features for such motions. Our current solution is to approximate
this with a region of interest in front of the hand. In image (a) we can
see the depiction of a reaching motion. We have in the scene three
objects which are supported by a surface. A gripper, representing the
human hand, and the region of interest in front of the hand. Whenever an
object overlaps this region of interest, it will be marked as a potential
object-to-be-reached, the occurrence timestamp will be saved, 
and its relative distance to the hand will be tracked.
In the depicted image the three objects are all overlapping the 
region of interest of the hand, where the blue stars represent the
moments when the objects are marked as a reach-to candidate.

In the next image (b) we showcase two main happenings: firstly,
({\textbullet}) the start of a potential \emph{fixation} action,
directly causing the end of the related reaching motion (still potential).
Fixation represents in our system: the movement of the hand 
--while in contact with the to-be-grasped object-- to the desired grasp
pose, however, not every reaching motion needs to be preceded by a
fixation. In the image the contact between the hand and the potential
object is shown by the small yellow ellipsoid area. Secondly,
({\textbullet}) we drawn the trajectory of the hand from its previous
pose until its current one. The trajectory is split into two colors:
a green one, representing the reaching trajectory, and a gray one which
would have been the theoretical starting point for a reaching motion,
if the previously mentioned rule would not had been broken, namely:
"the motion of the user's hand \emph{towards} an object". At one point
the hand was moving away from the objects, thus the start-to reach
time of the tracked objects was being constantly reset until the hand 
started moving back towards the targets again. In the image we can notice
the apple is no longer in the region of interest, however the 
milk carton is is. The model is still keeping track of the objects
in the region since it cannot know the intention of the user,
it might not grasp the object.

In image (c) the user grasped the object (grasp contact points 
shown in red). During the grasp the region of interest in no
longer required, and the cached potential objects are removed.
The hand trajectory is extended (shown in blue), representing the hand
fixation motion pattern. After grasping the object there are two possible
motions that can follow: (d) \emph{slide} or (e) \emph{pick-up}.
We define a sliding motion: a voluntary motion of a grasped object
on its supporting surface. The sliding motion in the example frame ended
when the object contact with its supporting plane ended. This could mean
that the object was either picked-up or directly transported. In the
example image, the object goes directly in the transport state. The
sliding trajectory is depicted in purple. The other possible case is a
pick-up motion, defined as: lifting a grasped object off its supporting
surface. Similarly as in the reach motion, pick-up does not have any
distinct features to segment it. We know it starts when the contact with
the supporting surface is broken, but cannot tell when it ends. 
For example, when picking up an object off the floor and placing it
on the top cupboard, it is hard to tell when the pick-up motion becomes 
a transport motion. For this we wen with a similar approach to reaching,
when the contact with the supporting surface is broken, a region of 
interest is created, and when the grasped object leaves this area,
the pick-up motion is marked as finished. In the image the trajectory is
depicted in yellow, and the leaving of the region of interest with a
purple star.

The following motion is \emph{transport}, and it is depicted in figure (f)
and (h). This is defined in our grammar as: the motion of a grasped 
object after a slide or pick-up, until either released, or a put-down
or a sliding motion occurs. In (g) we can observe a the transport motion
followed by directly by sliding. This happens if the grasped object is
not approaching the supporting surface from above. If the object
\emph{is} approaching the supporting surface from above,
a \emph{put-down} motion will occur. We know this ends when the object is
in contact with the supporting surface, however, we do not know when
did it start. For this reason during the transport motions we cache the
movements of the grasped object with a decay time, we illustrated this in
the images using a green transparent line beneath the hand trajectory.
The approach now becomes, whenever a put-down motion happens after the
transport, a region of interest area is created and we backtrack the
motions of the grasped object until it stops overlapping this area, we
then mark this as a start point of the put-down action. We depicted this 
situation in image (i). Similarly as in the the reaching case, if during
the put-down motion, the object is moving away from the supporting
surface, the put-down motion is cut short, starting only when there was
a continuous approaching motion towards the target. This case is depicted
in image (j).

\section{Experiments}
\label{sec:experiments}

\begin{figure}[htb]
    \begin{center}		
        \includegraphics[width=\columnwidth]%
        {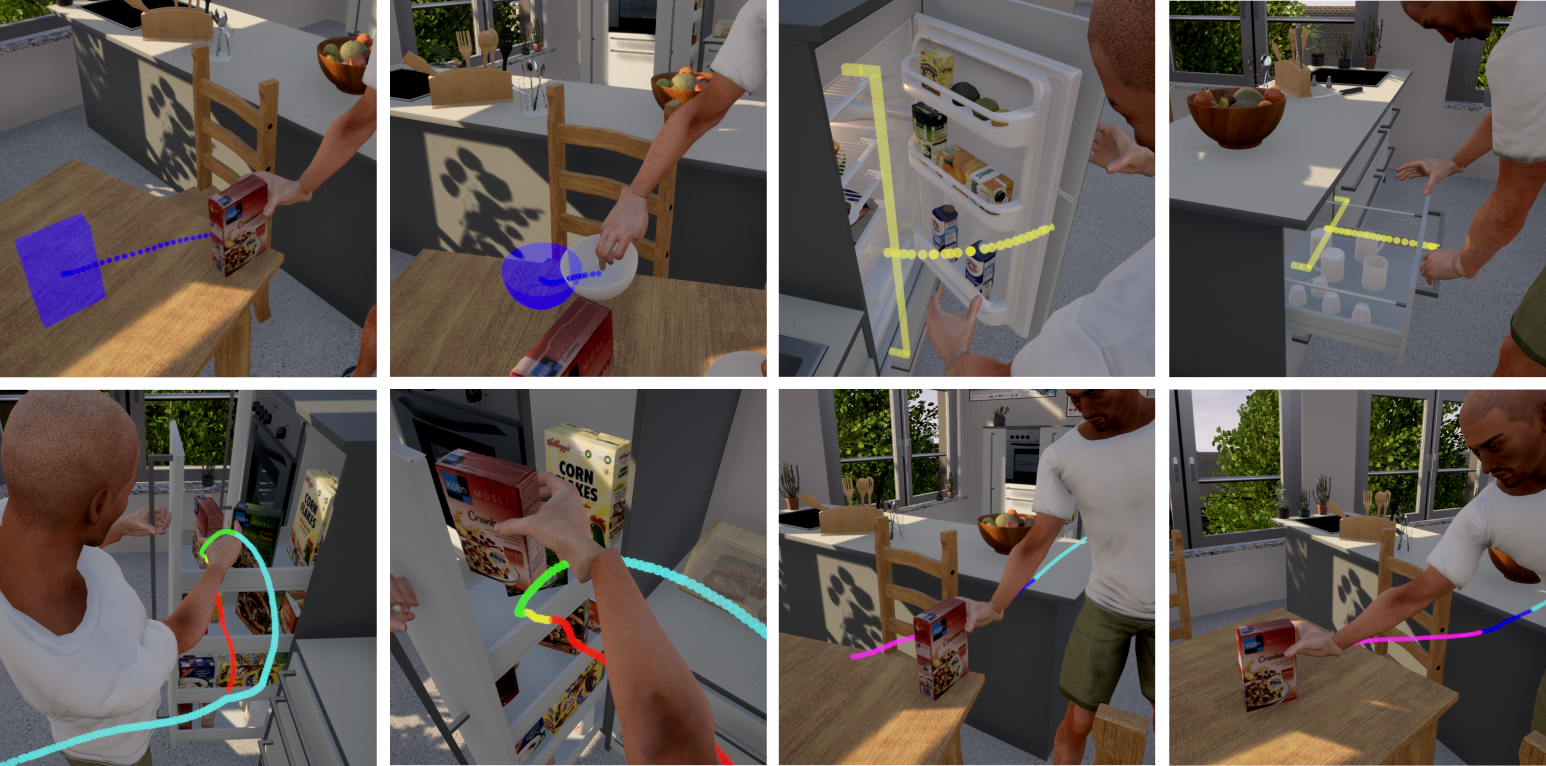}
        \caption{Query results}
        \vspace*{-0.5cm}
        \label{fig:qa} 
    \end{center}
\end{figure}

In the following section we briefly present a few shortened \knowrob\ 
\cite{beetz18knowrob} query examples on how specific data can be 
accessed and visualized from the hybrid knowledge base, the results
are visualized in Figure~\ref{fig:qa}. 

In the first example we
ask for directly detected events, without having any specific reasoning.
The query boils down to searching for a given action type acted on a given
object type. After a result is found, the action timestamps 
and the object instance are used to reconstruct the world at the given time
and append the queried markers to it. The results can be seen in the
top left part of the figure.

\begin{lstlisting}[style=Prolog]
   ?- entity(Act, [an, action, [type, 'Sliding'],
       [object, [an, object, [name, Obj], [type, 'CerealBox']]]]),
      occurs(Act, [Start, End]),
      show(world(Start)),
      show(trajectory(Obj), [Start, End]),
      show(marker(Obj), Start).
\end{lstlisting}

In the following query we take advantage of the partonomy representation 
of the world in order to narrow down our search for specific objects.
In this case we are querying for grasping handles belonging for specific
classes. We can se the results visualized in the top right side 
of the image.

\begin{lstlisting}[style=Prolog]
   ?- entity(Act, [an, action, [type, 'Grasping'],
        [object, [an, object, [name, Obj], [type, 'Handle']]]]),
      entity(Door, [an, object, [type, 'FridgeDoor']]),
      holds(part_of(Obj, Door)),
      occurs(Act, [Start, End]),
      show(world(Start)),
      show(trajectory(Obj), [Start, End]),
      show(marker(Obj), Start).
\end{lstlisting}

In the final query we are using sub-actions to define an upper action type.
In our case a pick-and-place action, which is not directly detected by 
the system, only their subparts. Rules are defined on which
sub-actions belong to the type, and their order can be specified using 
temporal reasoning \cite{allen}.

\begin{lstlisting}[style=Prolog]
   ?- entity(Act, [an, action, [type, 'PickAndPlace'],
       [object, [an, object, [name, Obj], [type, 'CerealBox']]],
       [sub-action, [an, action, [type, 'Reaching'], .. ]],
       [sub-action, [an, action, [type, 'Grasping'], .. ]],
       [..], 
       [sub-action, [an, action, [type, 'Sliding'], .. ]]),
      occurs(Act, [Start, End]),
      show(world(Start)),
      show(trajectory(Obj), [Start, End]), [..].
\end{lstlisting}

\section{Related Work}
\label{sec:related-work}

Similar to our approach, in \cite{vrgrasp19} the authors
present a similar system with the scope of collecting automatically 
annotated synthetic data form virtual environments, focusing mainly
on visual realism. In comparison the human interaction in our system is 
physics based, is capable of automatically detecting and annotating
actions and events in a robot understandable manner.

In \cite{uhde20} the authors use a dataset of human activities executed
in a virtual reality environment to generate hypotheses about casual
dependencies between actions. A consolidation of the two system would 
allow the possibility to collect and access labeled human activities
without the need of manual annotation.

\section{Conclusion and Future Work}
\label{sec:conclusion}

In this work we presented a framework able to automatically collect fully
annotated motion data from virtual environments. The data is stored in
a hybrid knowledge base, combining symbolic and subsymbolic representation,
understandable and accessible by robots. For the virtual environment
representation we are using Unreal Engine 4 with its underlying physics
engine NVIDIA PhysX.

As future research, we are planning to implement various improvements
to the system, such as avoiding one directional physics on the animated 
human body, integrating a refined physics interaction similar to the 
hand controllers. As a first step would be to use blended animations,
which has the capability to follow the tracked body animation and
to react physical interference. We would also like to use human
gaze patterns in order to refine the action detection heuristics, 
especially in cases where there are no other cues to use. Another
objective is to collect a large scale dataset similar to epic kitchen
~\cite{Damen2018EPICKITCHENS} and have it available online for robots
and researchers via the \openease\cite{tenorth15openease} platform.

\vspace{-1ex}
\bibliography{references}
\bibliographystyle{IEEEtran}
\end{document}